\documentclass[10pt,twocolumn,letterpaper]{article}

\usepackage{iccv}
\usepackage{times}
\usepackage{epsfig}
\usepackage{graphicx}
\usepackage{amsmath}
\usepackage{amssymb}
\usepackage{booktabs}

\usepackage[breaklinks=true,bookmarks=false]{hyperref}

\iccvfinalcopy % *** Uncomment this line for the final submission

\def\iccvPaperID{****} % *** Enter the ICCV Paper ID here

\ificcvfinal\fi

\begin{document}

%%%%%%%%% TITLE
\title{Skip-Connected Neural Networks with Layout Graphs for \newline Floor Plan Auto-Generation}

\author{Yuntae Jeon\\
Sungkyunkwan University\\
Suwon, Korea\\
{\tt\small jyt0131@g.skku.edu}
\and
Dai Quoc Tran\\
Sungkyunkwan University\\
Suwon, Korea\\
{\tt\small daitran@skku.edu}
\and
Seunghee Park\\
Sungkyunkwan University\\
Suwon, Korea\\
{\tt\small shparkpc@skku.edu}
}

\maketitle
\ificcvfinal\thispagestyle{empty}\fi

%%%%%%%%% ABSTRACT
\begin{abstract}
   With the advent of AI and computer vision techniques, the quest for automated and efficient floor plan designs has gained momentum. This paper presents a novel approach using skip-connected neural networks integrated with layout graphs. The skip-connected layers capture multi-scale floor plan information, and the encoder-decoder networks with GNN facilitate pixel-level probability-based generation. Validated on the MSD dataset, our approach achieved a 93.9 mIoU score in the 1st CVAAD workshop challenge. Code and pre-trained models are publicly available at \href{https://github.com/yuntaeJ/SkipNet-FloorPlanGen}{https://github.com/yuntaeJ/SkipNet-FloorPlanGen}.
\end{abstract}

% introduction, method, and results/discussion

%%%%%%%%% BODY TEXT
\section{Introduction}

Floor Plan auto-generation refers to the use of computational algorithms and tools to automatically design and optimize the spatial layout of a building or structure. Traditional floor plan design often requires substantial time, expertise, and manual iteration to balance both functional needs and aesthetic considerations. The auto-generation of floor plans offers a solution to this challenge by providing rapid, objective-driven designs that can maximize space utilization, enhance occupant comfort, and reduce design overhead.

% 따라서 많은 연구들이 이루어 졌지만, 확장성 문제 이번 대회에서는 large scale 그리고 class 기반 그래프를 주어준다.
In recent years, numerous studies have been conducted on floor plan auto-generation based on computer vision and deep learning. RPLAN \cite{wu2019data} suggests encoder-decoder networks for locating room and constructs 80k floor plans dataset from real residential buildings. Graph2Plan \cite{hu2020graph2plan} suggests graph neural networks(GNN) and convolutional neural networks(CNN) for graph-based floor plan generation using RPLAN dataset. There also exist GAN-based study \cite{nauata2020house} with a bubble diagram for input. However, there are still challenges that are hard to solve such as: \textbf{1) Scalability Issue}: Almost recent studies have been limited by exclusively using the RPLAN dataset, which is comprised of residential floor plans. This poses a limitation when attempting to apply to buildings with different purposes, such as office buildings, and also proves challenging for larger scale buildings. \textbf{2) Graph Utilization Issue}: In the boundary-based approach like Graph2Plan, nodes in the graph can only be used if they are placed correctly inside the boundary. On the other hand, studies utilizing the graph as a bubble diagram offer too much freedom, rendering the use of boundaries as input unfeasible.

% 제목 및 간략하게 한줄 쓰고
% 문제에 맞게 contribution 각각 씀

%------------------------------------------------------------------------
\begin{figure}[t]
  \centering
  \includegraphics[width=0.95\linewidth]{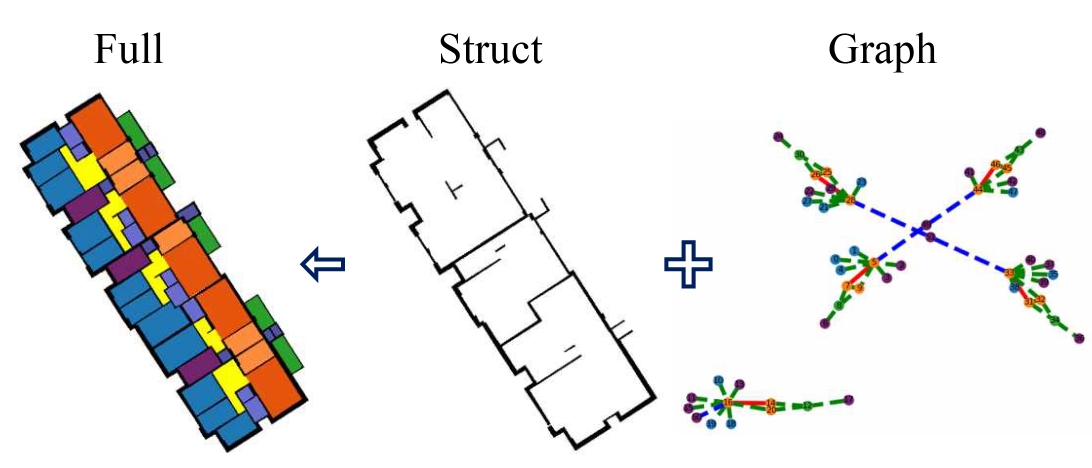}
   \caption{\textbf{Visualization} of floor plan auto-generation. The input is a struct(boundary info) and a graph(room types and connection), and the output is a generated floor plan called full.}
   \label{fig:overview}
\end{figure}
%-------------------------------------------------------------------------

We suggest encoder-decoder networks with skip connections for floor plan auto-generation. Our model inputs both a boundary image containing exterior information and a graph looks like bubble diagram showed in Fig. \ref{fig:overview}. We tested on Modified Swiss Dwellings (MSD) dataset \cite{swiss_dwelling_dataset} provided by 1st Computer Vision Aided Architectural Design (CVAAD) workshop on ICCV 2023. Our main contributions can be summarized as follows:

%------------------------------------------------------------------------
\begin{figure*}[t]
  \centering
  \includegraphics[width=0.9\linewidth]{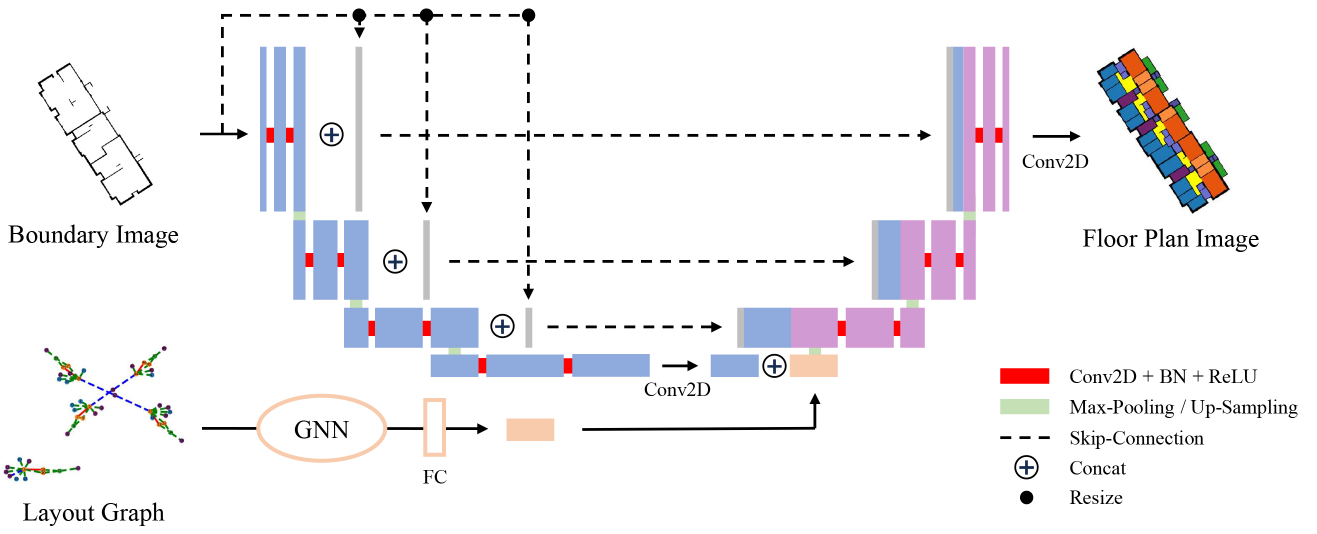}
   \caption{\textbf{Architecture} of our proposed SkipNet-FloorPlanGen}
   \label{fig:trajectory_prediction}
\end{figure*}
%-------------------------------------------------------------------------

\begin{enumerate}
    \item We utilized skip-connected layers to better comprehend various scale information of floor plans and validated this approach on the MSD dataset, which contains a diverse range of scales.
    \item We inferred bubble diagram-style graphs using GNN and concatenated the acquired graph features prior to the upsampling phase, enabling floor plan generation based on pixel-level probabilities.
\end{enumerate}

%------------------------------------------------------------------------
\section{Method}

\subsection{Boundary Image Pre-Processing}

Our pre-processing of boundary image begins by applying Mobile-SAM \cite{mobile_sam}, which is segment anything model for mobile. We generate segmentation masks and prioritize the largest one, to get the exterior part of the building structure. After that, we can structure a processed image composed of three channels: 'in-wall-out', taking a value of 1 for the interior, 0.5 for the boundary and 0 for the exterior; 'in-out', excluding wall information from previous channel; and the 'raw-boundary'. This structure is inspired by RPLAN \cite{wu2019data} dataset.

\subsection{Skip-Connected Neural Networks}

Our model employs a skip-connected architecture designed to preserve spatial details across various scales. The architecture comprises two central components: the encoder and the decoder, both supplemented with skip connections to ensure information flow across layers. The encoder component plays a role in extracting features from the input boundary image. Through a series of convolutional layers, it progressively down-samples the input while concurrently amplifying its feature dimensionality. This process enables the network to capture intricate patterns and semantics from the image at various scales. However, as the spatial dimensions are reduced, the risk of losing granular details increases.

The decoder acts as the counterbalance to the encoder. Tasked with the up-sampling of the condensed feature maps, the decoder employs skip-connections that bridge layers together. These connections reintroduce the lost spatial details from the encoding phase by directly linking the outputs of the encoder's layers to the decoder. In a strategic enhancement, our design also fuses the resized input boundary image at each decoding step. This novel integration ensures the generated floor plans are not just detailed but also strictly adhere to the input boundary constraints, ensuring the fidelity and accuracy of the generated outputs.

The combined effect of this encoder-decoder architecture, when fortified by the skip-connections, results in a more accurate and detail-preserving output. The network is equipped to understand and maintain the input boundary constraints efficiently across different scales, leading to enhanced consistency and fidelity in the generated floor plans.

\subsection{Graph Neural Networks}

We captures layout graph constraints using GNN, to ensure functionally feasible floor plans. We employ GCNConv layers for node representation learning, refining and aggregating the features to produce a 2D feature map. These graph features are then concatenated with the deepest outputs of the encoder, intertwining spatial details with layout graph constraints. As this merged data proceeds through the decoding process, the model seamlessly integrates both the spatial and topological information, yielding a floor plan that effectively combines visual precision with architectural layout constraints. 

%------------------------------------------------------------------------
\section{Results \& Discussion}

%------------------------------------------------------------------------
\begin{figure*}[t]
  \centering
  \includegraphics[width=0.7\linewidth]{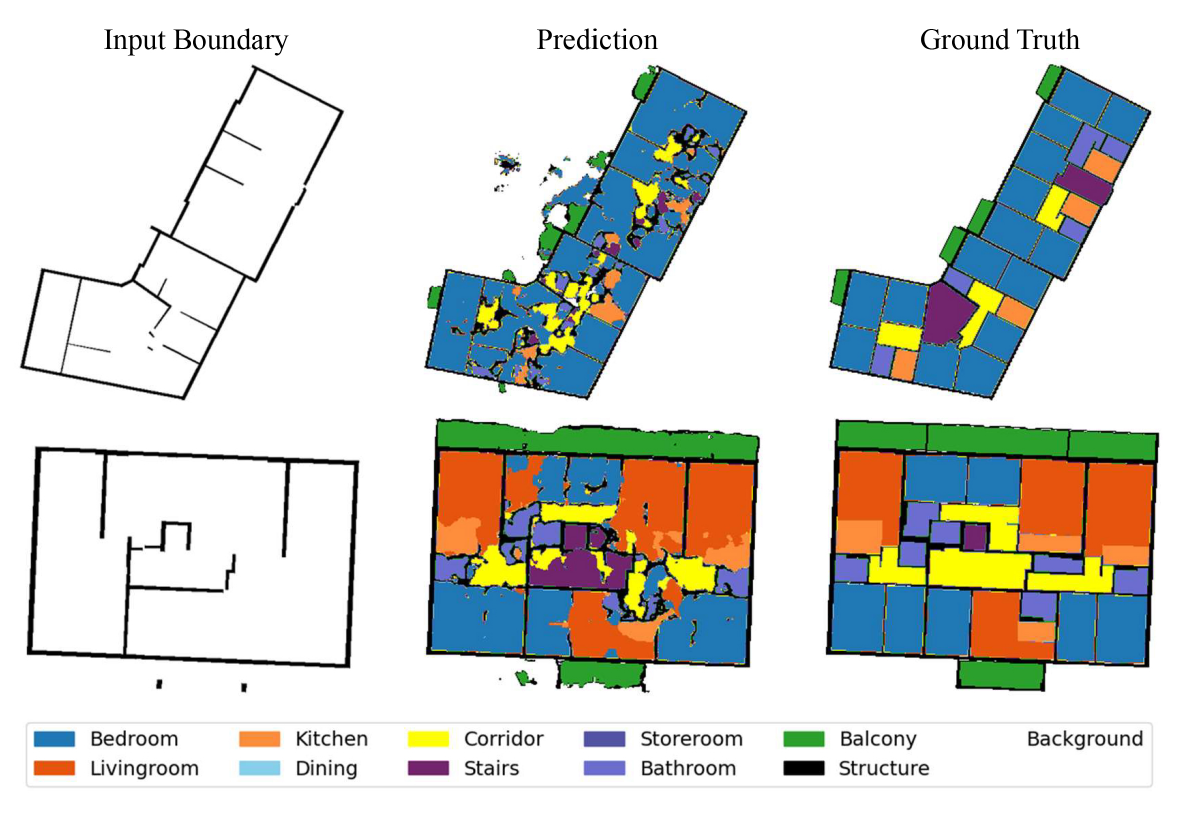}
   \caption{\textbf{Visualization} of floor plan auto-generation result.}
   \label{fig:result_vis}
\end{figure*}
%-------------------------------------------------------------------------

% experiment with implementation
1st CVAAD workshop at ICCV 2023 provided MSD dataset \cite{swiss_dwelling_dataset}, which includes boundary images, layout graphs and ground truth floor plans of single-as well as multi-unit building complexes across Switzerland, with 4167 floor plans for training and 1390 for testing. We will evaluate our model using Intersection over Union (IoU) that calculates the average intersection over union of predicted and ground truth segments across all classes. The training and inference processes were conducted on one NVIDIA A6000 GPU with PyTorch 2.0.0.

\subsection{Quantitative \& Qualitative Results}

%------------------------------------------------------------------------
\begin{table}[htbp]
    \caption{The results of 1st CVAAD workshop challenge in ICCV 2023. Where B. and S. stand for background and structure for each}
    \centering
    \begin{tabular}{crrrr}
        \toprule
        User      & IoU w/ B.    & IoU w/o B.   & IoU S.  \\
        \midrule
        Team 1 (Ours)  & \textbf{0.939} & \textbf{0.355} & 0.574   \\
        Team 2     & 	0.925        & 0.295        & \textbf{0.666}  \\
        Team 3     & 	0.815        & 0.222        & 0.406  \\
        \bottomrule
    \end{tabular}
    \label{tab:table_1}
\end{table}
%------------------------------------------------------------------------

Table \ref{tab:table_1} displays the competition leaderboard, demonstrating that the encoder-decoder model, enhanced with skip-connections and concatenation with resized boundary images, is a robust method. Fig. \ref{fig:result_vis} shows the qualitative result of our method. We separated a validation dataset from the train set, and Fig. \ref{fig:result_vis} illustrates the visualization results on the validation set.

\subsection{Discussion}

This paper presents a novel approach using skip-connected neural networks integrated with layout graphs. The skip-connected layers capture multi-scale floor plan information, and the encoder-decoder networks with GNN facilitate pixel-level probability-based generation with layout constraints. Our proposed method has been evaluated on the 1st CVAAD workshop MSD dataset \cite{swiss_dwelling_dataset} on ICCV 2023, and demonstrated its robust. 

% 부족한점 향후 연구할 방향 추가필요
In the future, we will focus on transforming boundary images into graph diagrams or vectorized forms for enhanced deep learning applications. The transition could mitigate limitations in scalable representations. Additionally, we aim to construct hierarchical or probabilistic graphs considering inter-room characteristics in layout graphs, aiming to pioneer a novel approach in handling spatial representations for more robust and scalable model architectures.

{\small

% \bibliographystyle{final}
% \bibliography{final}
}

\end{document}